\documentclass[letterpaper]{article} 
\usepackage{aaai25}  
\usepackage{times}  
\usepackage{helvet}  
\usepackage{courier}  
\usepackage[hyphens]{url}  
\usepackage{graphicx} 
\usepackage{natbib}  
\usepackage{caption} 
\usepackage{algorithm}

\usepackage{newfloat}
\usepackage{listings}

\usepackage{amssymb}
\usepackage{multirow}
\usepackage{algpseudocode}
\usepackage{subfigure} 
\usepackage{amsmath}
\usepackage{amsthm}
\usepackage{amsfonts}
\usepackage{color}
\usepackage{url}
\usepackage{etoolbox}
\usepackage{booktabs}
\usepackage{array}
\usepackage{dsfont}

\newtheorem{theorem}{Theorem}
\newtheorem{lemma}{Lemma}
\newtheorem{definition}{Definition}

\nocopyright

\DeclareCaptionStyle{ruled}{labelfont=normalfont,labelsep=colon,strut=off} 
\floatstyle{ruled}
\newfloat{listing}{tb}{lst}{}
\floatname{listing}{Listing}
\title{Private and Communication-Efficient Federated Learning based on Differentially Private Sketches}
\author{
    Meifan Zhang,Zhanhong Xie,Lihua Yin{\rm *}
}
\affiliations{

    Cyberspace Institute of Advanced Technology, Guangzhou University, Guangzhou, 510006, China
    
    zhangmf@gzhu.edu.cn, xzh@e.gzhu.edu.cn, yinlh@gzhu.edu.cn

    {\rm *}Lihua Yin is the corresponding author
}

\usepackage{bibentry}

\begin{document}

\maketitle

\begin{abstract}
Federated learning (FL) faces two primary challenges: the risk of privacy leakage due to parameter sharing and communication inefficiencies. To address these challenges, we propose DPSFL, a federated learning method that utilizes differentially private sketches. DPSFL compresses the local gradients of each client using a count sketch, thereby improving communication efficiency, while adding noise to the sketches to ensure differential privacy (DP). We provide a theoretical analysis of privacy and convergence for the proposed method. Gradient clipping is essential in DP learning to limit sensitivity and constrain the addition of noise. However, clipping introduces bias into the gradients, negatively impacting FL performance. To mitigate the impact of clipping, we propose an enhanced method, DPSFL-AC, which employs an adaptive clipping strategy. Experimental comparisons with existing techniques demonstrate the superiority of our methods concerning privacy preservation, communication efficiency, and model accuracy.
\end{abstract}

%

\section{Introduction}\label{sec:introduction}

Federated learning (FL) is a distributed machine learning approach that ensures data privacy and enables collaborative model updates by training models across multiple devices. Its core principle is to utilize the local computational resources of participants to train and update models without exchanging raw data. Recently, due to its exceptional privacy protection in decentralized environments, FL has emerged as a crucial technique in distributed machine learning~\cite{10.1145/3298981, 10.1093/jamia/ocaa341}, recommendation systems~\cite{Yang2020}, and medical applications~\cite{10.1145/3412357}.

Although FL avoids transmitting raw data, the sharing of local models still poses a risk of \textbf{privacy leakage}. Furthermore, the decentralized nature of FL introduces significant \textbf{communication costs} between clients and server.
First, the global model in FL is susceptible to membership inference attacks and model inversion attacks, where local models can be inferred from the global model, thereby threatening client privacy~\cite{Yeom_Giacomelli_Fredrikson_Jha_2018, Shokri_Stronati_Song_Shmatikov_2017}.
Second, the communication costs become substantial when training large model containing millions of parameters, even though each client communicates only updates or gradients with the server instead of raw data~\cite{10314794, 10255314, 10438925}.

To safeguard the privacy of each client in FL, differential privacy (DP)~\cite{10.1007/11787006_1} is a promising privacy-preserving tool, which injects noise into the parameters or gradients of the local model before transmission. Many studies adopt DP to ensure local data privacy in FL~\cite{abadi2016deep, DBLP:conf/iclr/McMahanRT018, wang2023lds}.
In FL with DP, clients typically perform two steps: gradient clipping and noise addition. Gradient clipping 
limits the gradient norm to a constant ($C$) to control sensitivity, adjusting gradient $g$ to $g/\max(1,\frac{{\|g\|}}{C})$, where sensitivity dictates the maximum impact of a single input change on aggregation. In the noise addition step, Laplace or Gaussian noise is added to the local stochastic gradient, proportional to sensitivity.While both steps can introduce errors that reduce model accuracy, especially for models with a large number of parameters~\cite{el2022differential}, safeguarding privacy while maintaining utility is challenging~\cite{DBLP:conf/iclr/TramerB21}. Smaller models can mitigate DP noise effects, but larger models are often necessary for better generalization in complex tasks~\cite{DBLP:conf/nips/BrownMRSKDNSSAA20}.

To reduce communication costs, several methods have been proposed, including \textit{quantization}~\cite{DBLP:conf/nips/TangGZZL18}, \textit{sparsification}~\cite{DBLP:conf/iclr/LinHM0D18}, \textit{incremental learning}~\cite{Dong_2022_CVPR}, and \textit{sketch-based compression}~\cite{pmlr-v202-gascon23a}.
\textit{Quantization} compresses gradient values into smaller representations, but it cannot support high compression ratio since the space cost is still linear to the dimension of gradient.
\textit{Sparsification} methods transmit only the \textit{top-k} significant gradients, but it fails to feedback the information loss in FL since each client participates only once.
\textit{Incremental learning} only transmits updates instead of the entire parameter set, potentially affecting performance and convergence rates.
Most of these methods lack convergence guarantee. Recent works~\cite{Ivkin_Rothchild_Ullah_Braverman_Stoica_Arora_2019, rothchild2020fetchsgd} have adopted sketch-based compression to capture compression errors while ensuring convergence, summarizing large datasets into a compact \textit{Count Sketch}~\cite{Charikar_Chen_Farach-Colton_2004}. While these methods partially address the communication issue, few of them jointly address privacy leakage. 

In summary, previous research has addressed privacy concerns and communication costs independently. However, achieving both privacy and communication efficiency in FL remains challenging and requires further investigation.

\textbf{Challenge 1: Reducing DP noise while enhancing communication efficiency.}
A straightforward approach to combining privacy preservation and gradient compression involves adding noise to the gradients to ensure DP and then sketching the noisy gradients to decrease communication costs. However, this combination of errors may adversely affect the convergence and accuracy of FL.
\textbf{Solution:} Instead of directly adding noise to the high-dimensional gradients, we propose adding DP noise to the compact sketch that summarizes the gradients. Since the noise error scales with the number of dimensions, sketch-based compression has the potential to reduce both communication costs and the amount of noise required to maintain privacy.

\textbf{Challenge 2: Reducing the impact of gradient clipping.}
As the sketch reduces the required amount of noise, clipping becomes the primary source of error.
A smaller clipping threshold affects a larger quantile of the gradients, requiring less DP noise, as it limits sensitivity to a smaller value. Conversely, a larger threshold results in a reduced clipping impact but necessitates more DP noise to maintain the desired level of privacy.
\textbf{Solution:} We propose an adaptive clipping strategy that dynamically adjusts the clipping threshold to limit the clipping impact on the \textit{top-k} updates, which primarily influence convergence. To privately collect and aggregate the local clipping impact, we also add DP noise to this local information before transmission; however, this incurs only a negligible fraction of the privacy budget.

\begin{figure}
	\centering
	\includegraphics[width=0.5\textwidth]{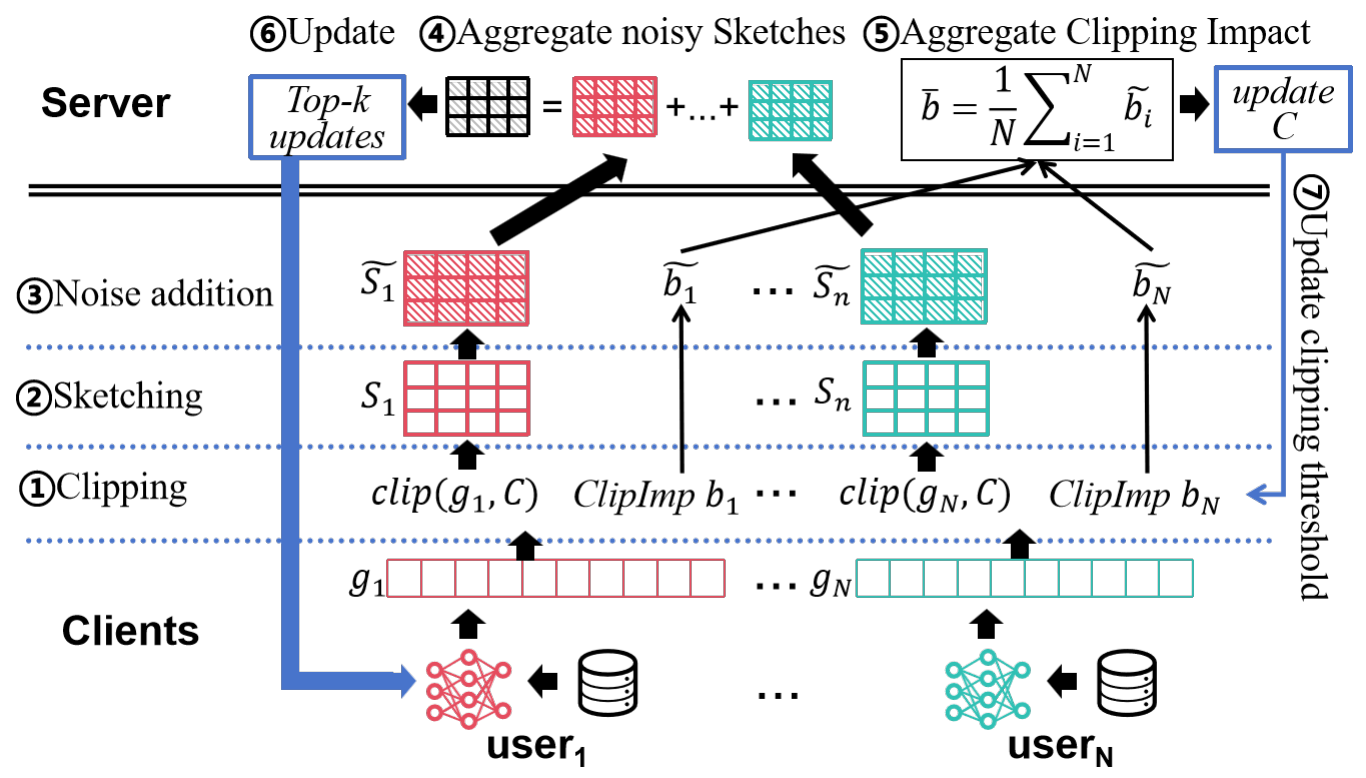}
	\caption{Overview of the DPSFL-AC framework.}
	\label{Fig1:The overview of our proposed scheme}
\end{figure}

To enhance the communication efficiency and reduce the DP noise, we propose a federated learning method based on differentially private sketches (DPSFL). To further reduce the clipping impact, we propose an enhanced method DPSFL-AC using an Adaptive Clipping strategy.
\textbf{Figure~\ref{Fig1:The overview of our proposed scheme} shows the workflow of our method.}
\textbf{Each client:}
\textcircled{1} clipping: train the local model, clip the gradient with a threshold $C$, and compute the impact of clipping;
\textcircled{2} sketching: use \textit{Count Sketch} to compress gradients;
\textcircled{3} noise addition: add noise to each counter of the sketch to ensure DP.
\textbf{The Server:}
\textcircled{4} aggregates the local noisy sketches;
\textcircled{5} aggregates the local impact of clipping;
\textcircled{6} finds the \textit{top-k} updates based on the aggregated sketch and broadcasts the update to the clients;
\textcircled{7} updates the clipping threshold $C$ to reduce the distance between the estimated and target clipping impact.

\textbf{Our main contributions}:
\begin{itemize}
	\item We propose a private and communication-efficient federated learning method based on differentially private sketches (\textbf{DPSFL}) and provide a theoretical analysis of both privacy and convergence.
	\item To further mitigate the impact of gradient clipping, we introduce an enhanced method, \textbf{DPSFL-AC}, which employs an \textbf{adaptive clipping} strategy.
	\item Experimental results demonstrate that the superiority of our methods in terms of privacy preservation, model performance, and communication efficiency.
\end{itemize}

\section{Related work}\label{sec:related_work}

\subsection{Differentially Private Federated Learning (DP+FL)}
In this paper, we assume that the server is honest but curious, indicating that local gradients or updates require privacy preservation prior to transmission.

\noindent\textbf{Noise addition.} As extensions of DPSGD~\cite{abadi2016deep}, DP+FL approaches add independent random noise to the local gradients or updates to avoid leaking the participant of local data~\cite{ijcai2021p217}.
Gaussian and Laplace mechanisms are widely used in DP+FL due to their simple mathematical privacy analysis~\cite{10360319}.
Skellam mechanism ~\cite{DBLP:journals/pvldb/BaoZXYOTA22} is proposed to overcome the non-closed noise of Gaussian mechanism under summation. FLAME~\cite{DBLP:conf/aaai/Liu0CGY21} uses a shuffle model to enhance the utility, but it requires a semi-trusted shuffler between clients and server.

\noindent\textbf{Clipping.} Clipping is a strategy utilized to limit sensitivity by bounding the norm of local gradients or updates. \cite{DBLP:conf/icml/ZhangCH0Y22} studies the impact of clipping on FL and 
concludes that a more concentrated distribution of client updates results in  smaller clipping bias, while a more dispersed distribution leads to larger bias.
\cite{DBLP:conf/sp/XiaoXWD23} emphasizes the influence of gradient clipping on the utility loss in DPSGD, and proposes using inner-outer momentum and BatchClipping to reduce bias.
To reduce the impact of clipping, DP-FedAvg~\cite{NEURIPS2021_91cff01a} adjusts the clipping threshold to a specified quantile of the update norm distribution. DP-PSAC~\cite{DBLP:conf/aaai/XiaSYF0XF23} uses non-monotonic adaptive clipping to ensure privacy without hyperparameter tuning, demonstrating superiority in vision and language tasks. \cite{DBLP:conf/ccs/Xiao0D23} introduces twice sampling and hybrid clipping techniques to enhance privacy and efficiency in DP deep learning.

\noindent\textbf{In summary}, DP is a promising privacy-preserving method for FL. However, most existing methods directly add noise to gradients, resulting in bringing significant noise to models with a large number of parameters. Furthermore, no previous work has studied the impact of combining DP noise and  clipping errors during the compression of local gradients.

\subsection{Communication-efficient Federated Learning}
Sketch-based compression is more promising to reduce the communication cost in distributed learning than traditional quantization and sparsification since it loses less useful local information thus it can achieve better convergence performance~\cite{DBLP:conf/icde/GuiSWHH23}. So we focus on sketch-based FL or distributed learning. 
SketchML~\cite{DBLP:conf/sigmod/JiangFY018} accelerates the distributed machine learning based on sketches, but it provides no convergence guarantee.
FetchSGD~\cite{rothchild2020fetchsgd} accumulates the error caused by \textit{top-k} operation to restore the convergence.
A recent work~\cite{pmlr-v202-gascon23a} employs invertible Bloom lookup tables in conjunction with subsampling to compute federated heavy hitter recovery, but it remains unclear how to extend this approach to FL.
SK-Gradient~\cite{DBLP:conf/icde/GuiSWHH23} uses index table to replace the hash computations to improve the computational efficiency.

\noindent\textbf{In summary}, sketch-based compression can effectively reduce communication costs with minimal information loss. However, some methods are designed for distributed learning rather than specifically for FL, and most existing techniques do not jointly address privacy leakage.

\section{Preliminaries}\label{sec:preliminaries}

\subsection{Differential Privacy}\label{Sec:Local Differential Privacy}

\begin{definition}
	\textbf{Differential privacy}~\cite{10.1007/11787006_1}
	A randomized mechanism $\mathcal{A}$ is 
	$(\epsilon,\delta)$-differentially private, if for any two adjacent datasets $D$, $D'$ that differ by only one record, and for any subset $S$ of possible outputs of $\mathcal{A}$, we have
	\begin{equation}
		{\Pr[\mathcal{A}({D}) \in S]} \le {e^\epsilon\cdot \Pr[\mathcal{A}({D'}) \in S] + \delta},
	\end{equation}
\end{definition}
\noindent where $\epsilon$ is the privacy budget, and $\delta$ is the failure probability.

\begin{definition}
	\textbf{Sensitivity.}
	Given a function $f$ and any two adjacent datasets $D$ and $D'$, the $l_2$ sensitivity of $f$ is denoted as $\Delta_f = \max_{D, D'} \| f(D) - f(D') \|$.
\end{definition}

To simplify the privacy analysis, we adopt the $\rho$-Zero Concentrated Differential Privacy ($\rho$-zCDP)~\cite{bun2016concentrated} instead of the original DP.
\begin{definition}
	\textbf{zCDP.}
	A randomized mechanism $\mathcal{A}$ satisfies $\rho$-Zero Concentrated Differential Privacy, if for any adjacent datasets $D,D'$ and all $\alpha\in (1,\infty)$,
	${D_\alpha }(\mathcal{A}(D)||\mathcal{A}(D')) \le \rho \alpha $,
	where $D_\alpha$ is the Renyi divergence.
\end{definition}

\begin{lemma}
	\textbf{Composition of zCDP.}
	Suppose $\mathcal{A}$ satisfies $\rho$- zCDP and $\mathcal{A}'$ satisfies $\rho'$-zCDP, and their composition   $\mathcal{A}''(D)=\mathcal{A}'(D,\mathcal{A}(D))$ satisfies $(\rho+\rho')$-zCDP.
\end{lemma}

\begin{lemma}\textbf{Transforming zCDP to DP.}
	If $\mathcal{A}$ satisfies $\rho$-zCDP, $\mathcal{A}$ satisfies $(\rho+2\sqrt{\rho\log(1/\delta )},\delta)$-DP.
\end{lemma}

\begin{theorem}
	\textbf{Privacy of Gaussian Mechanism.}
	Given a function $f$ with $l_2$-sensitivity $\Delta_2$, for any $\rho>0$, the Gaussian Mechanism defined as $\mathcal{A}(D)=f(D)+ \mathcal{N}(0,{\sigma ^2}{I_d})$ with noise level ${\sigma^2}=\Delta _2^2/(2\rho)$ satisfies $\rho$-zCDP, where $\mathcal{N}(0,{\sigma^2}{I_d})$ is a d-dimensional Gaussian random variable.
\end{theorem}

\subsection{FetchSGD~\cite{rothchild2020fetchsgd}}\label{Sec:FetchSGD}
\begin{figure}
	\centering
	\includegraphics[width=0.35\textwidth]{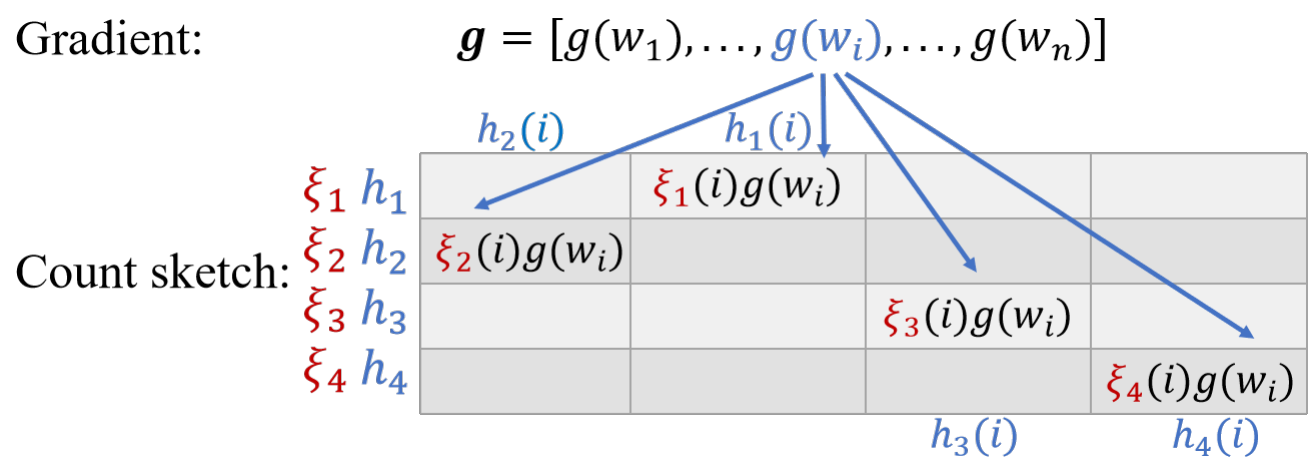}
	\caption{Gradient compression based on count sketch.}
	\label{countsketch}
\end{figure}

FetchSGD employs Count Sketch~\cite{Charikar_Chen_Farach-Colton_2004} to compress gradients in order to reduce communication overhead. 
\textbf{Count Sketch} is designed to estimate statistics such as frequency, heavy hitter, and inner product of massive data.
Its structure is a two-dim array with $l$ lines and $m$ columns as shown in Fig~\ref{countsketch}. Each line $j\in[l]$ has two corresponding hash functions $h_j: d\rightarrow m$ and $\xi_j:d\rightarrow \{-1,1\}$, where $d$ is the domain of key.

\noindent\textbf{Gradient Compression.} Inserting each coordinate $g(w_i)$ of a local gradient $g = [g(w_1),\ldots, g(w_n)]$ to the local sketch, aggregating all the local sketches, and estimating the \textit{top-k} heavy coordinates based on the aggregated sketch.

\noindent\textbf{Insert$(g,S)$}: for each $g(w_i)\in g$ and for each $j\in [l]$, $S[j, h_j(i)] += \xi_j(i) \cdot g(w_i)$.
In this way, each client $i$ compress its local gradients $g_i$ into a count sketch $S_i=S(g_i)$.

\noindent\textbf{Aggregate sketches}:  $S = \frac{1}{N}\sum_{i = 1}^N S_i$. The server aggregates the sketches instead of the gradients, leveraging the linearity of Count Sketch, i.e., $S(g_1+g_2) = S(g_1)+S(g_2)$.

\noindent\textbf{Estimation}: 
$\hat{g}(w_i)=\mathop{median}_{j\in[l]}\{\xi_j(i) \cdot S[j, h_j(i)]\}$.

\section{FL based on DP Sketches (DPSFL)}\label{sec:LDPS-FL}
	\subsection{The framework of DPSFL}
	DPSFL comprises client-side and server-side algorithms. \textbf{Each client} clips the local gradient, compresses it into a count sketch, and adds DP noise to the sketch before transmitting it to the server. \textbf{The server} aggregates all the noisy sketches, approximates the \textit{top-k} gradients, and broadcasts the updates to the clients. We provide the descriptions of some important notations in this paper in Table~\ref{table:notation}.
	\begin{table}[htbp]
			\centering
			\caption{Descriptions of some Frequent Notations}
			\label{table:notation}
			\begin{tabular}{lp{6cm}} 
					\toprule[1.5pt]
					\textbf{Symbol} & \textbf{Descriptions} \\
					\midrule[1pt]
					$(\epsilon,\delta)$  & Parameters in DP\\
					$\rho$  & Parameter in zCDP\\
					${\Delta_f}$  & The sensitivity of function $f$ \\
					$(\epsilon_S,\delta_S)$ & Parameters in count sketch\\
					$(l,m)$  & Rows and columns of a count sketch \\
					$\tau$ & Parameter for heavy coordinate\\
					$k$ & Parameter for $top\text{-}k$\\
					${w^t}$  & The model in the $t$th round \\
					$g_i^t$  & The gradient of client $i$ in the $t$th round\\
					$\beta$ & Momentum parameter\\
					$\eta$ & Learning rate\\
					$C$  & Gradient clipping threshold\\
					$(\theta,\gamma,\eta_C)$  & Parameters for adaptive clipping\\
					\bottomrule[1.5pt]
			\end{tabular}
	\end{table}

	\textbf{Client-side (Alg.~\ref{algorithm:Client-side LDPS-FL})}:  
	Each client $i$ of the $N$ selected clients in one round performs local model training in batches (line 2), clips the gradient (line 3), and compresses it using count sketch (line 4). It then calculates the sensitivity (line 5), adds Gaussian noise to the sketch according to the sensitivity to ensure DP (line 6), and sends the noisy sketch to the server (line 7).
	\begin{algorithm}[htbp]
		\caption{DPSFL-Client($i, {w^t}, \rho, C$)}\label{algorithm:Client-side LDPS-FL}
		\begin{algorithmic}[1]
			\State{Receive global model: $w_i^t \leftarrow {w^t}$}
			\State{Compute stochastic gradient $g_i^t$ on batch $B$:
			 $g_i^t = \frac{1}{|B|}\sum\nolimits_{j = 1}^{|B|} \nabla_w  L(w_i^t,{x_j},{y_j})$}
			\State{Clipping: $g_i^t = g_i^t/\max (1,\frac{{||g_i^t||}}{C})$}
			\State{Sketching: ${S_i} = S(g_i^t)$}
			\State{Sensitivity: $\Delta_S=C\sqrt{l}$}
			\State{Noise addition: $S_i = S_i + z_i$, $z_i \sim \mathcal{N}(0,\Delta_S^2/(2\rho))$}
			\State{Send $S_i$ to the server}
		\end{algorithmic}
	\end{algorithm}

	\textbf{Server-side (Alg.~\ref{algorithm:Server-side LDPS-FL})}: It is the same with the server-side of \textit{FetchSGD}. First, the server initializes $S_u^0$ (for momentum accumulation of the aggregated sketches),
	$S_e^0$ (feeds back the error caused by \textit{top-k} operation), and the global model $w^0$ (lines 1-2). 
	The global model is sent to the selected clients in each round (lines 3-6), and the server aggregates the local sketches from the clients (line 7) to obtain a global gradient sketch (line 9). It introduces a momentum term to accelerate convergence and then adjusts error feedback (lines 10-11). It then selects the \textit{top-k} largest coordinates, updates the error feedback sketch $S_e^{t+1}$ and the global model (lines 12-14).
	
	\begin{algorithm}[htbp]
		\caption{DPSFL-Server}\label{algorithm:Server-side LDPS-FL}
		\hspace*{0.02in} {{\bf Input:} Sketch size $(l\times m)$, learning rate $\eta$, global iteration count $T$, momentum parameter $\beta$, number of selected clients $N$, privacy budget for zCDP $\rho$, clipping threshold $C$}\\
		\hspace*{0.02in} {{\bf Output:}  Final model ${w^T}$}
		\begin{algorithmic}[1]
			\State {Initialize $S_u^0$, $S_e^0$ to zero sketches}.
			\State {Initialize the global model ${w^0}$}. 
			\For{$t= 0, 1, \dots, T-1$}
				\State{Randomly select $N$ clients}
					\For{$i = 1,2, \dots, N$}
						\State {send the update of global model $w^t$ to client ${c_i}$}
						\State {$S _i^t$  $\leftarrow$ {\bf DPSFL-Client}($i, w^t, \rho, C$)}
					\EndFor
				\State{Aggregate sketches ${S^t} = \frac{1}{N}\sum_{i = 1}^N S_i^t$}
				\State{Momentum :${S_u^t = \beta S_u^{t - 1} + S^t}$}
				\State{Error feedback :${S_e^t = \eta S_u^t + S_e^t}$}
				\State{Unsketch:${{\Delta ^t} = Top\text{-}k(U(S_e^t))}$}
				\State{Error accumulation:${S_e^{t + 1} = S_e^t - S({\Delta ^t})}$}
				\State{Update ${{w ^{t + 1}} = {w^t} - {\Delta ^t}}$}
			\EndFor
			\State{return ${w^T}$}
		\end{algorithmic}
	\end{algorithm}

	\subsection{Privacy Analysis}
	DPSFL adds Gaussian noise to each counter of the sketch before transmission to protect the privacy.
	The required noise is determined by the sensitivity of the sketch, hence we first introduce the definition of sensitivity for the sketch.
	
	\begin{definition}\label{Def:sketch_sensitivity}
		The sensitivity $\Delta_S$ of a sketch $S$ (consisting of $l$ lines and $m$ columns) summarizing a vector of gradients $g=\{g_1, g_2, ..., g_n\}$ is defined as:
			$\Delta_S = \max_{g, g'} \|S(g) - S(g')\|_2$,
		where $g$ and $g'$ differ by at most one coordinate, i.e., $g[i]\ne g'[i]$ and $g[j]= g'[j]$ for all $j\ne i$.
	\end{definition}
	
	According to Definition~\ref{Def:sketch_sensitivity}, for each corresponding line of sketch  $S(g)$ and $S(g')$, they differ by at most one counter, and the difference is the distance between $g_i$ and $g'_i$. If each element $g_i$ is unbounded, $\Delta_S$ will be extremely large.
	
	\noindent\textbf{Limit the sensitivity by clipping.} Clipping is a common strategy in previous works~\cite{pmlr-v139-mai21a,pmlr-v202-koloskova23a} to limit the contributions of each gradient vector. We limit the norm of the gradient $g$ by $clip(g,C)=\frac{g}{\max(1,\frac{\|g\|}{C})}$, which ensures that $\|clip(g,C)\|_2\le C$. Therefore, $\|S(g)-S(g')\|_2=\sqrt{\sum_{j=1}^{l}(S(g)[j]-S(g')[j])^2}\le C\sqrt{l}$.
	According to Definition~\ref{Def:sketch_sensitivity}, the sensitivity $\Delta_S=C\sqrt{l}$.

	\begin{theorem}\label{Theorem:PrivacyAnalysis_zCDP}
		DPSFL with a gradient clipping threshold $C$ satisfies $\rho$-zCDP by adding independent Gaussian noises $\mathcal{N}(0, \sigma^2)$ to each counter of the sketch $S(l,m)$,
		where $\sigma = C\sqrt{\frac{l}{2\rho}}$. And it also satisfies $(\rho+2\sqrt{\rho\log(1/\delta )},\delta)$-DP.
	\end{theorem}

 \section{Convergence Analysis of DPSFL}
	In DPSFL, each client processes its local gradients in three steps before transmission: (1) \textbf{clipping} to reduce sensitivity, (2) \textbf{sketching} to minimize communication costs, and (3) \textbf{adding noise} to ensure DP. These steps collectively affect FL convergence, yet no prior studies have addressed convergence in this complex setting.
	We divide the convergence analysis into two steps: the convergence for sketch-based FL with clipping and sketching, excluding DP perturbation (DPSFL-NonNoise); and the convergence of DPSFL. \textbf{The proof of theorems and lemmas are in supplementary}.
	
	\subsection{Convergence of DPSFL-NonNoise}\label{section:Convergence_DPSFL-NonNoise}
	
	We analyze the convergence of DPSFL-NonNoise (Clipping + Sketching + NonNoise) by proving that the outputs $\{w^t\}_{t=1}^T$ have a guarantee that the upper bound of gradient size $\|\nabla f(w^t)\|^2$ goes to zero as $T\rightarrow \infty$.
	We first prove two lemmas before the proof of convergence.
	
	\noindent\textbf{Step 1. Reduce the update of DPSFL-NonNoise to an SGD-like update.}
		
	\begin{lemma}\label{lemma:ReduceUpdateToSGD}
		Let $\{w^t\}_{t=1}^{T}$ be the sequence of models generated by DPSFL-NonNoise with a gradient clipping threshold $C$.  
		Assume that the model $f$ is $L$-smooth and $\alpha\|g^t\|\le\|clip(g^t, C)\|$.
		Consider the virtual sequence $\tilde{w}^t=w^t-e^t-\frac{\eta\beta}{1-\beta} u^{t-1}$ with a step size $\eta$ and momentum parameter $\beta$, where $\{u^t\}_{t=1}^{T}$ and $\{e^t\}_{t=1}^{T}$ are the momentum and error accumulation vectors without sketching, we have 
		\begin{align}
			&\mathbb{E}[f(\tilde{w}^{t+1})]\!
			\le\!\mathbb{E}[f(\tilde{w}^{t})]-\frac{\eta}{1-\beta}(1-\frac{(\alpha-2)^2}{2})\|\nabla f({w}^{t})\|^2 \nonumber\\
			&+\frac{L\eta^2 C^2}{2(1-\beta)^2}
			+ \frac{L^2\eta}{1-\beta} \mathbb{E}\|e^t+\frac{\eta\beta}{1-\beta}u^{t-1}\|^2.
		\end{align}
	\end{lemma}
	
	Since both $e^t$ and $u^{t-1}$ are estimated based on sketches $S(e^t)$ and $S(u^{t-1})$ in DPSFL, we also need to estimate the bound of $\|e^t+\frac{\eta\beta}{1-\beta}u^{t-1}\|^2$ based on sketches.
	
	\noindent\textbf{Step 2. Compute the bound of $\|e^t+\frac{\eta\beta}{1-\beta}u^{t-1}\|^2$.}
	
	We first introduce the definition of heavy coordinate, and then review the error bound of count sketch in theorem~\ref{theorem:ErrorBoundCoundSketch}.
	\begin{definition}\label{Def:HeavyHitter}
		A gradient coordinate $g[i]$ is a $(\tau, l_2^2)$-heavy coordinate in a gradient vector $g$ if $g[i]^2\ge \tau\|g\|_2^2$.
	\end{definition}
	Count sketch can approximately find the heavy coordinates of a vector of gradients without clipping as follows.
	\begin{theorem}\label{theorem:ErrorBoundCoundSketch}
		A count sketch using $O(\frac{1}{\tau}\log(d/\delta_s))$ space, can find all $(\tau, l_2^2)$-heavy coordinates of a vector $g$, and approximate their values with probability at least $1-\delta_s$, where $d$ is the dimension of $g$.
	\end{theorem}
	According to this theorem, we can estimate the bound of $\|e^t+\frac{\eta\beta}{1-\beta}u^{t-1}\|^2$ based on count sketch in lemma~\ref{lemma:BoundofSketch}.
	
	
	\begin{lemma}\label{lemma:BoundofSketch}
		With a clipping threshold $C$ and a sketch in $O(\frac{1}{\tau}\log(d/\delta_{s}))$ space,  with probability at least $1-\delta_{s}$, 
		\begin{equation}
			\|e^t+\frac{\eta\beta}{1-\beta} u^{t-1}\|^2
			\!\le\!\frac{\eta^2C^2(1+\tau)^2}{(1-\beta)^2(1-\tau)^2} [\frac{2\beta^2}{(1-\beta)^2}+\frac{8(1-\tau)}{\tau^2}].\nonumber
		\end{equation}
	\end{lemma}

	\noindent\textbf{Step 3. The convergence of DPSFL-NonNoise}
	
	\begin{theorem}
		Let $\{w^t\}_{t=1}^{T}$ denote the sequence of models generated by DPSFL-NonNoise, and each local sketch requires $O(\frac{1}{\tau}\log\frac{d\cdot T}{\delta_S})$ space. Assume that the model $f$ is $L$-smooth, with a clipping threshold $C$ and $\alpha \|g^t\|\le \|clip(g^t, C)\|$.	If the step size is set to $\eta=\frac{1-\beta}{L\sqrt{T}}$, we have
		\begin{equation}
			\min_{t=1\dots T}\mathbb{E}\|\nabla f(w^t)\|^2
			\le\frac{L\cdot D_f+C^2}{b\sqrt{T}}+\frac{v(1+\tau)^2C^2}{b(1-\tau)^2 T},
		\end{equation}
		$D_f=f(w^0)-f^*$, $b=1-\frac{(\alpha-2)^2}{2}$, $v=\frac{2\beta^2}{(1-\beta)^2}+\frac{8(1-\tau)}{\tau^2}$.
	\end{theorem}

	\subsection{Convergence of DPSFL}\label{section:Convergence_DPSFL}

	We first analyse the bound of noise, and then prove the convergence of DPSFL in the following three steps. 
	\begin{theorem}\label{theorem:DPSFL-NonNoise_Convergence}
		\textbf(Bound of noise.)
		Suppose that we insert independent Gaussian noise sampled from $\mathcal{N}(0,\sigma^2)$ to each counter of a count sketch $S$ containing $(\frac{1}{\tau}\log\frac{1}{\delta_S})$ counters to satisfy $\rho$-zCDP, where $\sigma = C\sqrt{\frac{\log(1/\delta_S)}{2\rho}}$. 
		Let $z_{i,j}$ be the noise added to one counter, with the probability at least $1-\delta_S$ that for any $i\in[\log(1/\delta_S)]$ and $j\in[1/\tau]$, we have  
		   $|z_{i,j}|\le C\sqrt{\frac{\log(1/\delta_S)}{\rho}}\cdot \sqrt{\log\frac{2\log(1/\delta_S)}{\delta_S}}$.
	\end{theorem}

	\noindent\textbf{Step 1. Analyze the update of DPSFL.}
	\begin{lemma}\label{lemma:updata_DPSFL}
		Let $\{w^t\}_{t=1}^{T}$ be the sequence of models generated by DPSFL with a gradient clipping threshold $C$. The noise added to each counter of a sketch containing $(\frac{1}{\tau}\log\frac{1}{\delta_S})$ counters is sampled from $\mathcal{N}(0,\sigma^2)$. With other conditions same with those in lemma~\ref{lemma:ReduceUpdateToSGD}, we have 
		 
		\begin{align}
			&\mathbb{E}[f(\tilde{w}^{t+1})]\!
			\le\!\mathbb{E}[f(\tilde{w}^{t})]\!-\!\frac{\eta}{1-\beta}(1-\frac{(\alpha-2)^2}{2})\|\nabla(f({w}^{t}))\|^2 \nonumber \\
		 &\!\!+\!\frac{L\eta^2 C^2(1+{NoiImp})}{2(1-\beta)^2}\!+\!\frac{L^2\eta}{1\!\!-\!\!\beta}\mathbb{E}\|e^t\!+\!\frac{\eta\beta}{1\!-\!\beta}u^{t-1}\!\|^2,
		\end{align}
		where $NoiImp=\frac{1}{\rho\cdot\tau}\log\frac{1}{\delta_S}\log\frac{\frac{2}{\tau}\log\frac{1}{\delta_S}}{\delta_S}$ indicates the \textit{Impact of Noise}.
	\end{lemma}
	
	\noindent\textbf{Step 2. Estimate $\|e^t\|^2$ and $\|u^{t}\|^2$ based on noisy sketch.}
	In the remainder of this section, the $NoiImp$ value is consistent with that in Lemma~\ref{lemma:updata_DPSFL}.
	With a noisy sketch $\tilde{S}=S+z$ containing $l\times m$ counters, where $m=\frac{1}{\tau}$, and $l=\log\frac{1}{\delta_S}$, and $z=\mathcal{N}(0,{\sigma^2}{I_{l,m}})$ is a $(l\times m)$-dim Gaussian noise.
	$\left|\|\tilde{S}(x)\|-\|x\|\right|\le \tau\|x\|+C^2\cdot NoiImp$.
	
	\begin{lemma}\label{lemma:BoundofNoisySketch}
		With a clipping threshold $C$ and a sketch costing {$O(\frac{\log(d/\delta_S)}{(\sqrt{\tau}-2\sqrt{\text{NoiImp}})^2})$ space}, with probability at least $1-\delta_{S}$, $\|e^t+\frac{\eta\beta}{1-\beta} u^{t-1}\|^2 \le\frac{\eta^2 C^2(1+\text{NoiImp})(1+\tau)^2}{(1-\beta)^2(1-\tau)^2} [\frac{2\beta^2}{(1-\beta)^2}+\frac{8(1-\tau)}{\tau^2}]$.
	\end{lemma}

	\noindent\textbf{Step 3. The convergence of DPSFL.}
	
	\begin{theorem}\label{theorem:DPSFL_Convergence}
		 Let $\{w^t\}_{t=1}^{T}$ denote the sequence of models generated by DPSFL, and each local sketch requires $O\left(\frac{\log(dT/\delta_S)}{(\sqrt{\tau}-2\sqrt{\text{NoiImp}})^2}\right)$ space and incorporates Gaussian noise sampled from $\mathcal{N}(0,\sigma^2)$ in each counter of the sketch. Assume that the model $f$ is $L$-smooth, with a clipping threshold $C$, and that $\alpha \|g^t\|\le\| \text{clip}(g^t, C)\|$. If the step size is set to $\eta = \frac{1-\beta}{L\sqrt{T}}$, then
		\begin{align}
			&\min_{t=1\dots T}\mathbb{E}\|\nabla f(w^t)\|^2
			\le\!\frac{L(f(w^0)-f^*)+C^2(1+{NoiImp})}{b\sqrt{T}}\nonumber\\
			&+\frac{v(1+\tau)^2 C^2(1+{NoiImp})}{b(1-\tau)^2 T},
		\end{align}
		$D_f=f(w^0)-f^*$, $b=1-\frac{(\alpha-2)^2}{2}$, $v=\frac{2\beta^2}{(1-\beta)^2}+\frac{8(1-\tau)}{\tau^2}$.
	\end{theorem}
	
	A comparison of Theorem~\ref{theorem:DPSFL_Convergence} and Theorem~\ref{theorem:DPSFL-NonNoise_Convergence} reveals that inserting DP noise necessitates an increase in the required size of the local sketch and delays the convergence.

 \section{DPSFL-AC: DPSFL with Adaptive Clipping}\label{sec:ADPS-FL}
	
\begin{figure}[htbp]
		\centering
		\includegraphics[width=0.26\textwidth]{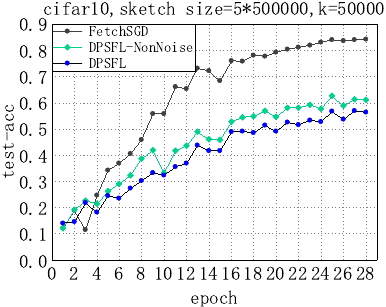}
		\caption{The impact of clipping}
		\label{fetchclipping}
	\end{figure}

	DPSFL introduces both clipping errors and noise errors, with the latter mitigated through sketch-based compression. This section focuses on reducing the impact of clipping. As illustrated in Fig.~\ref{fetchclipping}, the gap between FetchSGD (without clipping and DP noise) and DPSFL-NonNoise (with clipping) is significantly larger than the gap between DPSFL-NonNoise and DPSFL (with both clipping and DP noise). Therefore, the gap between FetchSGD and DPSFL is partly due to DP noise but primarily due to gradient clipping.

	However, selecting an appropriate gradient clipping threshold is challenging. A low threshold may exclude valuable gradient information, introducing significant bias, while a high threshold can introduce excessive noise, potentially degrading model utility. To mitigate this issue, we propose an adaptive clipping strategy that focuses on limiting the impact of clipping to the \textit{top-k} gradient coordinates that directly influence the model updates.

	\subsection{Adaptive Clipping}
	We dynamically adjust the clipping threshold $C$ by reducing the distance between the estimated probability and a given probability $\gamma$ that the impact of clipping on the \textit{top-k} coordinates can be bounded with a threshold $\theta$. 
	\begin{equation}
	\Pr[\|Top(clip(g,C))-Top(g)\|\le \theta \|Top(g)\|]=\gamma,
	\end{equation}
	where $\|Top(g)\|$ is the $l_2$ norm of \textit{top-k} coordinates of $g$, and $\|Top(clip(g,C))-Top(g)\|$ is the clipping error of \textit{top-k} coordinates. Note that, since each client participating in the $t$th round is unaware of the \textit{top-k} coordinates of the global model in the current round in FL setting, we utilize the \textit{top-k} coordinate indices from the $(t-1)$th round to measure the impact of clipping. As the server broadcasts the \textit{top-k} sparse updates to some selected clients in each round, it is straightforward for the clients to obtain the indices along with the \textit{top-k} updates.

	\subsection{Private aggregation of the clipping impact}
	To safeguard privacy, each client computes the impact of clipping on its local gradient and adds DP noise before transmitting it to the server. The server then aggregates the local clipping impacts and uses the global clipping impact to adjust the clipping threshold.
	
	\textbf{Client-side (Alg.~\ref{algorithm:Client-side ADPS-FL}):}
	Each client $i$ uses a single bit $b_i$ to denote whether it satisfies $\|Top(clip(g,C))-Top(g)\|\le\theta\|Top(g)\|$ or not (line 2). If the condition holds, $b_i=1$, otherwise, $b_i=0$. To ensure DP, it adds a Gaussian noise $z_b$ to $b_i$ before transmission (line 3). Finally, each client sends the noisy bit along with the noisy sketch to the server.

	\begin{algorithm}
	\caption{DPSFL-AC (Client)}\label{algorithm:Client-side ADPS-FL}
		\begin{algorithmic}[1]
		\State Line 1-6 of the Alg.1.
		\State $b_i=\mathds{1}\{\|Top(clip(g_i,C))-Top(g_i)\|\le \theta \|Top(g_i)\|\}$
		\State $\tilde{b}_i=b_i+ z_b$, $z_b \sim N(0,{\sigma_b^2})$
		\State{\textbf{return:} $(S_i, \tilde{b}_i)$}
		\end{algorithmic}
	\end{algorithm}
	
	\textbf{Server-side (Alg.~\ref{algorithm:Server-side ADPS-FL}):} 
	The server aggregates each noisy bit $\tilde{b}_i$ from each client and computes the average (line 2). It then adjusts the clipping threshold with a learning rate $\eta_C$ to reduce the distance between the estimated probability $\overline{b}^t$ and the target probability $\gamma$ (line 3).
	
	\begin{algorithm}
	\caption{DPSFL-AC (Server)}\label{algorithm:Server-side ADPS-FL}
		\begin{algorithmic}[1]
		\State{Add the following two lines to Alg.2 after line 14.}
		\State{Aggregate the clipping impact: $\overline{b}^t = \frac{1}{N}\sum_{i = 1}^N \tilde{b}_i^t$}
		\State{Update clipping threshold:$C\leftarrow C\cdot \exp(-\eta_C(\overline{b}^t-\gamma))$}
		\end{algorithmic}
	\end{algorithm}
	As the clipping impact is denoted as a single bit, it requires only a negligible fraction of the privacy budget.

 \section{Experiments}\label{sec:experiments}
\subsection{Experimental Settings}\label{Sec:Experimental Settings}
\noindent \underline{\textbf{Hardware}}
A Ubuntu (20.04.1) machine equipped with 8 GeForce RTX 2080 Ti GPUs.

\noindent \underline{\textbf{Datasets}}
We conduct the experiments on three image classification datasets, \textbf{CIFAR10}, \textbf{CIFAR100}~\cite{CIFAR}, and  \textbf{MNIST}\cite{DBLP:journals/spm/Deng12}.

\noindent \underline{\textbf{Model description}}

\textbf{ResNet9}: Model for experiments on CIFAR10, CIFAR100 and MNIST. It contains approximately 7 million parameters, and the detail of its structure is in the {supplementary.}

\noindent \underline{\textbf{Comparison method}}

(1) \textbf{FedAvg}~\cite{pmlr-v54-mcmahan17a}: The baseline federated learning method, that each client sends the local gradients to the server \textbf{without adding DP noise}. 

(2) \textbf{FetchSGD}~\cite{rothchild2020fetchsgd}: It compresses gradients using Count Sketch \textbf{without adding DP noise}.

(3) \textbf{DPFL}: The combination of DPSGD~\cite{abadi2016deep} and FedAvg~\cite{pmlr-v54-mcmahan17a} by adding DP noise to the local gradients before transmission. 

(4) \textbf{Adaptive-DPFL (ADPFL)}~\cite{10262055}: The client uses global model parameters and local historical gradients to adjust the sensitivity and adaptively add noise.

\noindent \underline{\textbf{Measuring index}}

(1) \textbf{Accuracy} ($acc$): the ratio of the number of samples correctly predicted by the model (True Positives) to the total number of samples (Total). 
$acc = \frac{\rm{TruePositive}}{\rm{Total}} \times 100\%$

(2) \textbf{Compression Level} ($CL$): the compression level relative to uncompressed SGD in terms of total bytes uploaded and downloaded. 
$CL = \frac{\rm{Baseline}}{\rm{ComCost}}$, where we set the \textit{Baseline} as 1.6 million Mib and the \textit{ComCost} stands for the true communication cost.

\noindent \underline{\textbf{Other default settings}}.
\textbf{Parameters for DP:}  $\varepsilon=4$ and $\delta=1e-5$;
\textbf{Parameters for clipping:} $C=1.5$, $\gamma=0.9$, $\theta=0.5$, $\sigma_b=0.1$, $\eta_C$=0.01;
\textbf{Parameters for sketch:} $(l\times m)=5\times500,000$, and $k=50000$ for top-$k$ updates.

\subsection{Experimental Results}\label{Sec:Experimental Results}
\begin{figure*}[htbp]
	\centering
	\subfigure[CIFAR10]{
		\centering
		\begin{minipage}[b]{0.28\textwidth}
			\includegraphics[width=1\textwidth,height=0.7\linewidth, trim={0cm 0cm 0cm 0cm}, clip]{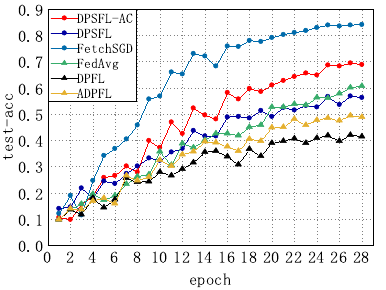}
		\end{minipage}
		\label{test-acc-cifar10}
	}
	\subfigure[CIFAR100]{
		\centering
		\begin{minipage}[b]{0.28\textwidth}
			\includegraphics[width=1\textwidth,height=0.7\linewidth, trim={0cm 0cm 0cm 0cm}, clip]{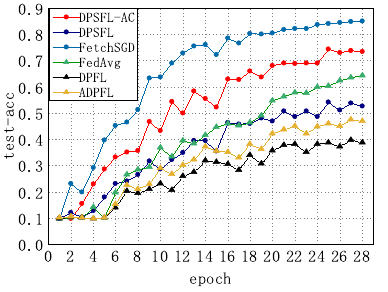}
		\end{minipage}
		\label{test-acc-cifar100}
	}
	\subfigure[MNIST]{
		\centering
		\begin{minipage}[b]{0.28\textwidth}
			\includegraphics[width=1\textwidth,height=0.7\linewidth, trim={0cm 0cm 0cm 0cm}, clip]{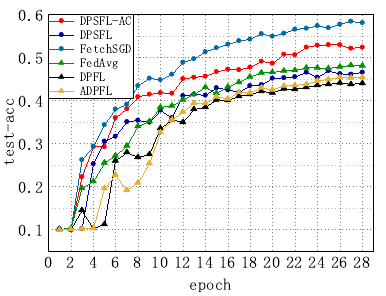}
		\end{minipage}
		\label{test-acc-mnist}
	}
	\caption{Comparison of Utility.}
	\label{cifar-mnist-acc}
\end{figure*}

\subsubsection{Utility Analysis (Figure~\ref{cifar-mnist-acc}).}

The accuracy of our DPSFL is comparable to the non-DP method FedAvg, and more accurate than other DP+FL methods, including DPFL and ADPFL. Our enhanced method DPSFL-AC demonstrates significantly greater accuracy than others except the non-DP method FetchSGD. 
DPSFL outperforms DPFL and ADPFL by adding noise to the compact sketch rather than the high-dimensional gradient. The distance between DPSFL and FetchSGD, along with the similarity between DPSFL and FedAvg, confirms that clipping is the primary source of error in DPSFL. Furthermore, DPSFL-AC is closer to FetchSGD since it reduces the impact of clipping.

\begin{figure*}[htbp]
	\centering
	\subfigure[CIFAR10]{
		\centering
		\begin{minipage}[b]{0.28\textwidth}
			\includegraphics[width=1\textwidth,height=0.7\linewidth, trim={0cm 0cm 0cm 0cm}, clip]{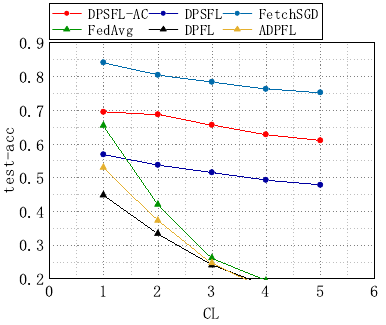}
		\end{minipage}
		\label{compression-cifar10}
	}
	\subfigure[CIFAR100]{
		\centering
		\begin{minipage}[b]{0.28\textwidth}
			\includegraphics[width=1\textwidth,height=0.7\linewidth, trim={0cm 0cm 0cm 0cm}, clip]{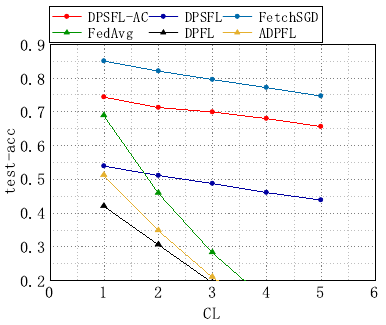}
		\end{minipage}
		\label{compression-cifar100}
	}
	\subfigure[MNIST]{
		\centering
		\begin{minipage}[b]{0.28\textwidth}
			\includegraphics[width=1\textwidth,height=0.7\linewidth, trim={0cm 0cm 0cm 0cm}, clip]{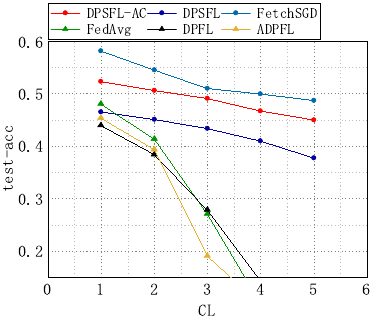}
		\end{minipage}
		\label{compression-mnist}
	}
	\caption{Comparison of communication efficiency.}
	\label{cifar-mnist-communication}
\end{figure*}

\subsubsection{Communication Efficiency (Figure~\ref{cifar-mnist-communication}).}
We test the accuracy of different methods with compression level ($CL$) varing from 1 to 5. To adjust the $CL$ for sketch-based methods, we try values of $k\in [50, 32, 25, 18, 12]\times 10^3$ for \textit{top-k} updates and $m\in [500, 300, 200, 120, 80]\times 10^3$ for the number of columns. 
Methods adopting no sketches such as FedAvg achieve the same $CL$ by reducing the number of iterations.
If $CL=1$ (no compression), the accuracy of FedAvg outperforms DPSFL. As the $CL$ increases, the accuracy decreases by only about 0.1 for the methods using the count sketch (lines with dots), whereas it declines rapidly for the methods not using the count sketch (lines with triangles).
The main reason is that count sketch effectively compress gradients without much information loss, allowing DPSFL-AC, DPSFL, and FetchSGD to perform more iterations under the same communication cost, thereby improving the accuracy. In contrast, under the same communication cost, FedAvg, DPFL, and ADPFL have fewer training iterations, leading to insufficient model training and a subsequent drop in accuracy.

\begin{figure}[htbp]
	\centering
	\begin{minipage}[t]{0.49\linewidth}
		\centering
		\includegraphics[width=1\textwidth, trim={0cm 0cm 0cm 0cm}, clip]{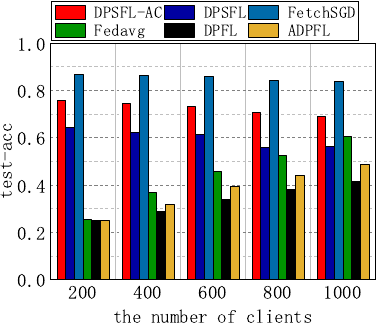}
		\caption{The impact of number of clients (CIFAR10)}
		\label{num_clients}
	\end{minipage}
	\begin{minipage}[t]{0.49\linewidth}
		\centering
		\includegraphics[width=1\textwidth, trim={0cm 0cm 0cm 0cm}, clip]{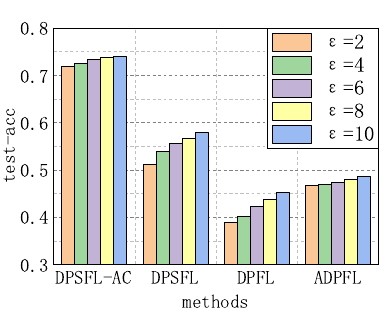}
		\caption{The impact of privacy budget (CIFAR100)}
		\label{epsilon-acc}
	\end{minipage}
\end{figure}

\subsubsection{Impact of number of clients (Figure~\ref{num_clients}).}
We evaluate the impact of the number of clients by conducting experiments on CIFAR10. 
As the number of clients increases, the accuracy of DPSFL-AC, DPSFL, and FetchSGD slightly decreases, while the accuracy of FedAvg, DPFL, and ADPFL gradually increases.
{Sketch-based compression leads to some information loss, which is amplified as the number of clients increases. As more clients join, the diversity of data distribution increases, and the model requires more information to accurately update its parameters. }

\subsubsection{Impact of privacy budget (Figure~\ref{epsilon-acc}).}
We set the privacy budget $\epsilon\in\{2, 4, 6, 8, 10\}$, and conduct experiments on CIFAR100. 
The results show that as $\epsilon$ increases, the accuracy of all methods improves. 
The impact of $\epsilon$ varies on different methods. DPSFL-AC and ADPFL, which employ adaptive clipping, show the least sensitivity to changes in $\epsilon$, maintaining relatively high accuracy even at lower privacy budgets. This suggests that adaptive clipping effectively mitigates some of the negative effects of noise addition by better controlling the sensitivity of gradients, thereby reducing the overall impact of noise on the learning process. Conversely, methods without such adaptive mechanisms show more improvements as $\epsilon$ increases, highlighting their vulnerability to the trade-offs of DP.

\section{Conclusion}\label{sec:conclusion}
This paper proposes two sketch-based methods for FL under DP. The first method, DPSFL, compresses gradients using count sketch to achieve high communication efficiency and inserts noise to the sketch to ensure DP. The second method, DPSFL-AC, enhances accuracy by adaptively adjusting the clipping threshold. Future work will focus on further mitigating the impact of clipping on FL with DP sketches.

\bibliography{aaai25}

\begin{thebibliography}{41}
\providecommand{\natexlab}[1]{#1}

\bibitem[{Abadi et~al.(2016)Abadi, Chu, Goodfellow, McMahan, Mironov, Talwar, and Zhang}]{abadi2016deep}
Abadi, M.; Chu, A.; Goodfellow, I.; McMahan, H.~B.; Mironov, I.; Talwar, K.; and Zhang, L. 2016.
\newblock Deep learning with differential privacy.
\newblock In \emph{Proceedings of the 2016 ACM SIGSAC conference on computer and communications security}, 308--318.

\bibitem[{Andrew et~al.(2021)Andrew, Thakkar, McMahan, and Ramaswamy}]{NEURIPS2021_91cff01a}
Andrew, G.; Thakkar, O.; McMahan, B.; and Ramaswamy, S. 2021.
\newblock Differentially Private Learning with Adaptive Clipping.
\newblock In Ranzato, M.; Beygelzimer, A.; Dauphin, Y.; Liang, P.; and Vaughan, J.~W., eds., \emph{Advances in Neural Information Processing Systems}, volume~34, 17455--17466. Curran Associates, Inc.

\bibitem[{Bao et~al.(2022)Bao, Zhu, Xiao, Yang, Ooi, Tan, and Aung}]{DBLP:journals/pvldb/BaoZXYOTA22}
Bao, E.; Zhu, Y.; Xiao, X.; Yang, Y.; Ooi, B.~C.; Tan, B. H.~M.; and Aung, K. M.~M. 2022.
\newblock Skellam Mixture Mechanism: a Novel Approach to Federated Learning with Differential Privacy.
\newblock \emph{Proc. {VLDB} Endow.}, 15(11): 2348--2360.

\bibitem[{Brown et~al.(2020)Brown, Mann, Ryder, Subbiah, Kaplan, Dhariwal, Neelakantan, Shyam, Sastry, Askell, Agarwal, Herbert{-}Voss, Krueger, Henighan, Child, Ramesh, Ziegler, Wu, Winter, Hesse, Chen, Sigler, Litwin, Gray, Chess, Clark, Berner, McCandlish, Radford, Sutskever, and Amodei}]{DBLP:conf/nips/BrownMRSKDNSSAA20}
Brown, T.~B.; Mann, B.; Ryder, N.; Subbiah, M.; Kaplan, J.; Dhariwal, P.; Neelakantan, A.; Shyam, P.; Sastry, G.; Askell, A.; Agarwal, S.; Herbert{-}Voss, A.; Krueger, G.; Henighan, T.; Child, R.; Ramesh, A.; Ziegler, D.~M.; Wu, J.; Winter, C.; Hesse, C.; Chen, M.; Sigler, E.; Litwin, M.; Gray, S.; Chess, B.; Clark, J.; Berner, C.; McCandlish, S.; Radford, A.; Sutskever, I.; and Amodei, D. 2020.
\newblock Language Models are Few-Shot Learners.
\newblock In Larochelle, H.; Ranzato, M.; Hadsell, R.; Balcan, M.; and Lin, H., eds., \emph{Advances in Neural Information Processing Systems 33: Annual Conference on Neural Information Processing Systems 2020, NeurIPS 2020, December 6-12, 2020, virtual}.

\bibitem[{Bun and Steinke(2016)}]{bun2016concentrated}
Bun, M.; and Steinke, T. 2016.
\newblock Concentrated differential privacy: Simplifications, extensions, and lower bounds.
\newblock In \emph{Theory of Cryptography Conference}, 635--658. Springer.

\bibitem[{Charikar, Chen, and Farach-Colton(2004)}]{Charikar_Chen_Farach-Colton_2004}
Charikar, M.; Chen, K.; and Farach-Colton, M. 2004.
\newblock Finding frequent items in data streams.
\newblock \emph{Theoretical Computer Science}, 3–15.

\bibitem[{Cui et~al.(2024)Cui, Yang, Wu, Feng, and Hu}]{10314794}
Cui, Z.; Yang, T.; Wu, X.; Feng, H.; and Hu, B. 2024.
\newblock The Data Value Based Asynchronous Federated Learning for UAV Swarm Under Unstable Communication Scenarios.
\newblock \emph{IEEE Transactions on Mobile Computing}, 23(6): 7165--7179.

\bibitem[{Deng(2012)}]{DBLP:journals/spm/Deng12}
Deng, L. 2012.
\newblock The {MNIST} Database of Handwritten Digit Images for Machine Learning Research [Best of the Web].
\newblock \emph{{IEEE} Signal Process. Mag.}, 29(6): 141--142.

\bibitem[{Dong et~al.(2022)Dong, Wang, Fang, Sun, Xu, Wang, and Zhu}]{Dong_2022_CVPR}
Dong, J.; Wang, L.; Fang, Z.; Sun, G.; Xu, S.; Wang, X.; and Zhu, Q. 2022.
\newblock Federated Class-Incremental Learning.
\newblock In \emph{Proceedings of the IEEE/CVF Conference on Computer Vision and Pattern Recognition (CVPR)}, 10164--10173.

\bibitem[{Dwork(2006)}]{10.1007/11787006_1}
Dwork, C. 2006.
\newblock Differential Privacy.
\newblock In Bugliesi, M.; Preneel, B.; Sassone, V.; and Wegener, I., eds., \emph{Automata, Languages and Programming}, 1--12. Berlin, Heidelberg: Springer Berlin Heidelberg.
\newblock ISBN 978-3-540-35908-1.

\bibitem[{El~Ouadrhiri and Abdelhadi(2022)}]{el2022differential}
El~Ouadrhiri, A.; and Abdelhadi, A. 2022.
\newblock Differential privacy for deep and federated learning: A survey.
\newblock \emph{IEEE access}, 10: 22359--22380.

\bibitem[{Elmahallawy, Luo, and Ramadan(2024)}]{10438925}
Elmahallawy, M.; Luo, T.; and Ramadan, K. 2024.
\newblock Communication-Efficient Federated Learning for LEO Constellations Integrated With HAPs Using Hybrid NOMA-OFDM.
\newblock \emph{IEEE Journal on Selected Areas in Communications}, 42(5): 1097--1114.

\bibitem[{Gascon et~al.(2023)Gascon, Kairouz, Sun, and Suresh}]{pmlr-v202-gascon23a}
Gascon, A.; Kairouz, P.; Sun, Z.; and Suresh, A.~T. 2023.
\newblock Federated Heavy Hitter Recovery under Linear Sketching.
\newblock In Krause, A.; Brunskill, E.; Cho, K.; Engelhardt, B.; Sabato, S.; and Scarlett, J., eds., \emph{Proceedings of the 40th International Conference on Machine Learning}, volume 202 of \emph{Proceedings of Machine Learning Research}, 10997--11012. PMLR.

\bibitem[{Gui et~al.(2023)Gui, Song, Wang, He, and Huang}]{DBLP:conf/icde/GuiSWHH23}
Gui, J.; Song, Y.; Wang, Z.; He, C.; and Huang, Q. 2023.
\newblock SK-Gradient: Efficient Communication for Distributed Machine Learning with Data Sketch.
\newblock In \emph{39th {IEEE} International Conference on Data Engineering, {ICDE} 2023, Anaheim, CA, USA, April 3-7, 2023}, 2372--2385. {IEEE}.

\bibitem[{Hu, Guo, and Gong(2024)}]{10360319}
Hu, R.; Guo, Y.; and Gong, Y. 2024.
\newblock Federated Learning With Sparsified Model Perturbation: Improving Accuracy Under Client-Level Differential Privacy.
\newblock \emph{IEEE Transactions on Mobile Computing}, 23(8): 8242--8255.

\bibitem[{Ivkin et~al.(2019)Ivkin, Rothchild, Ullah, Braverman, Stoica, and Arora}]{Ivkin_Rothchild_Ullah_Braverman_Stoica_Arora_2019}
Ivkin, N.; Rothchild, D.; Ullah, E.; Braverman, V.; Stoica, I.; and Arora, R. 2019.
\newblock Communication-efficient Distributed SGD with Sketching.

\bibitem[{Jiang et~al.(2018)Jiang, Fu, Yang, and Cui}]{DBLP:conf/sigmod/JiangFY018}
Jiang, J.; Fu, F.; Yang, T.; and Cui, B. 2018.
\newblock SketchML: Accelerating Distributed Machine Learning with Data Sketches.
\newblock In Das, G.; Jermaine, C.~M.; and Bernstein, P.~A., eds., \emph{Proceedings of the 2018 International Conference on Management of Data, {SIGMOD} Conference 2018, Houston, TX, USA, June 10-15, 2018}, 1269--1284. {ACM}.

\bibitem[{Koloskova, Hendrikx, and Stich(2023)}]{pmlr-v202-koloskova23a}
Koloskova, A.; Hendrikx, H.; and Stich, S.~U. 2023.
\newblock Revisiting Gradient Clipping: Stochastic bias and tight convergence guarantees.
\newblock In Krause, A.; Brunskill, E.; Cho, K.; Engelhardt, B.; Sabato, S.; and Scarlett, J., eds., \emph{Proceedings of the 40th International Conference on Machine Learning}, volume 202 of \emph{Proceedings of Machine Learning Research}, 17343--17363. PMLR.

\bibitem[{Krizhevsky(2012)}]{CIFAR}
Krizhevsky, A. 2012.
\newblock Learning Multiple Layers of Features from Tiny Images.
\newblock \emph{University of Toronto}.

\bibitem[{Lin et~al.(2018)Lin, Han, Mao, Wang, and Dally}]{DBLP:conf/iclr/LinHM0D18}
Lin, Y.; Han, S.; Mao, H.; Wang, Y.; and Dally, B. 2018.
\newblock Deep Gradient Compression: Reducing the Communication Bandwidth for Distributed Training.
\newblock In \emph{6th International Conference on Learning Representations, {ICLR} 2018, Vancouver, BC, Canada, April 30 - May 3, 2018, Conference Track Proceedings}. OpenReview.net.

\bibitem[{Liu et~al.(2021)Liu, Cao, Chen, Guo, and Yoshikawa}]{DBLP:conf/aaai/Liu0CGY21}
Liu, R.; Cao, Y.; Chen, H.; Guo, R.; and Yoshikawa, M. 2021.
\newblock {FLAME:} Differentially Private Federated Learning in the Shuffle Model.
\newblock In \emph{Thirty-Fifth {AAAI} Conference on Artificial Intelligence, {AAAI} 2021, Thirty-Third Conference on Innovative Applications of Artificial Intelligence, {IAAI} 2021, The Eleventh Symposium on Educational Advances in Artificial Intelligence, {EAAI} 2021, Virtual Event, February 2-9, 2021}, 8688--8696. {AAAI} Press.

\bibitem[{Mai and Johansson(2021)}]{pmlr-v139-mai21a}
Mai, V.~V.; and Johansson, M. 2021.
\newblock Stability and Convergence of Stochastic Gradient Clipping: Beyond Lipschitz Continuity and Smoothness.
\newblock In Meila, M.; and Zhang, T., eds., \emph{Proceedings of the 38th International Conference on Machine Learning}, volume 139 of \emph{Proceedings of Machine Learning Research}, 7325--7335. PMLR.

\bibitem[{McMahan et~al.(2017)McMahan, Moore, Ramage, Hampson, and Arcas}]{pmlr-v54-mcmahan17a}
McMahan, B.; Moore, E.; Ramage, D.; Hampson, S.; and Arcas, B. A.~y. 2017.
\newblock {Communication-Efficient Learning of Deep Networks from Decentralized Data}.
\newblock In Singh, A.; and Zhu, J., eds., \emph{Proceedings of the 20th International Conference on Artificial Intelligence and Statistics}, volume~54 of \emph{Proceedings of Machine Learning Research}, 1273--1282. PMLR.

\bibitem[{McMahan et~al.(2018)McMahan, Ramage, Talwar, and Zhang}]{DBLP:conf/iclr/McMahanRT018}
McMahan, H.~B.; Ramage, D.; Talwar, K.; and Zhang, L. 2018.
\newblock Learning Differentially Private Recurrent Language Models.
\newblock In \emph{6th International Conference on Learning Representations, {ICLR} 2018, Vancouver, BC, Canada, April 30 - May 3, 2018, Conference Track Proceedings}. OpenReview.net.

\bibitem[{Pfitzner, Steckhan, and Arnrich(2021)}]{10.1145/3412357}
Pfitzner, B.; Steckhan, N.; and Arnrich, B. 2021.
\newblock Federated Learning in a Medical Context: A Systematic Literature Review.
\newblock \emph{ACM Trans. Internet Technol.}, 21(2).

\bibitem[{Rothchild et~al.(2020)Rothchild, Panda, Ullah, Ivkin, Stoica, Braverman, Gonzalez, and Arora}]{rothchild2020fetchsgd}
Rothchild, D.; Panda, A.; Ullah, E.; Ivkin, N.; Stoica, I.; Braverman, V.; Gonzalez, J.; and Arora, R. 2020.
\newblock Fetchsgd: Communication-efficient federated learning with sketching.
\newblock In \emph{International Conference on Machine Learning}, 8253--8265. PMLR.

\bibitem[{Sarma et~al.(2021)Sarma, Harmon, Sanford, Roth, Xu, Tetreault, Xu, Flores, Raman, Kulkarni, Wood, Choyke, Priester, Marks, Raman, Enzmann, Turkbey, Speier, and Arnold}]{10.1093/jamia/ocaa341}
Sarma, K.~V.; Harmon, S.; Sanford, T.; Roth, H.~R.; Xu, Z.; Tetreault, J.; Xu, D.; Flores, M.~G.; Raman, A.~G.; Kulkarni, R.; Wood, B.~J.; Choyke, P.~L.; Priester, A.~M.; Marks, L.~S.; Raman, S.~S.; Enzmann, D.; Turkbey, B.; Speier, W.; and Arnold, C.~W. 2021.
\newblock {Federated learning improves site performance in multicenter deep learning without data sharing}.
\newblock \emph{Journal of the American Medical Informatics Association}, 28(6): 1259--1264.

\bibitem[{Shokri et~al.(2017)Shokri, Stronati, Song, and Shmatikov}]{Shokri_Stronati_Song_Shmatikov_2017}
Shokri, R.; Stronati, M.; Song, C.; and Shmatikov, V. 2017.
\newblock Membership Inference Attacks Against Machine Learning Models.
\newblock In \emph{2017 IEEE Symposium on Security and Privacy (SP)}.

\bibitem[{Sun, Qian, and Chen(2021)}]{ijcai2021p217}
Sun, L.; Qian, J.; and Chen, X. 2021.
\newblock LDP-FL: Practical Private Aggregation in Federated Learning with Local Differential Privacy.
\newblock In Zhou, Z.-H., ed., \emph{Proceedings of the Thirtieth International Joint Conference on Artificial Intelligence, {IJCAI-21}}, 1571--1578. International Joint Conferences on Artificial Intelligence Organization.
\newblock Main Track.

\bibitem[{Tang et~al.(2018)Tang, Gan, Zhang, Zhang, and Liu}]{DBLP:conf/nips/TangGZZL18}
Tang, H.; Gan, S.; Zhang, C.; Zhang, T.; and Liu, J. 2018.
\newblock Communication Compression for Decentralized Training.
\newblock In Bengio, S.; Wallach, H.~M.; Larochelle, H.; Grauman, K.; Cesa{-}Bianchi, N.; and Garnett, R., eds., \emph{Advances in Neural Information Processing Systems 31: Annual Conference on Neural Information Processing Systems 2018, NeurIPS 2018, December 3-8, 2018, Montr{\'{e}}al, Canada}, 7663--7673.

\bibitem[{Tram{\`{e}}r and Boneh(2021)}]{DBLP:conf/iclr/TramerB21}
Tram{\`{e}}r, F.; and Boneh, D. 2021.
\newblock Differentially Private Learning Needs Better Features (or Much More Data).
\newblock In \emph{9th International Conference on Learning Representations, {ICLR} 2021, Virtual Event, Austria, May 3-7, 2021}. OpenReview.net.

\bibitem[{Wang et~al.(2023)Wang, Yang, Zhu, Wang, Su, and Sato}]{wang2023lds}
Wang, T.; Yang, Q.; Zhu, K.; Wang, J.; Su, C.; and Sato, K. 2023.
\newblock LDS-FL: Loss Differential Strategy based Federated Learning for Privacy Preserving.
\newblock \emph{IEEE Transactions on Information Forensics and Security}.

\bibitem[{Xia et~al.(2023)Xia, Shen, Yao, Fu, Xu, Xu, and Fu}]{DBLP:conf/aaai/XiaSYF0XF23}
Xia, T.; Shen, S.; Yao, S.; Fu, X.; Xu, K.; Xu, X.; and Fu, X. 2023.
\newblock Differentially Private Learning with Per-Sample Adaptive Clipping.
\newblock In Williams, B.; Chen, Y.; and Neville, J., eds., \emph{Thirty-Seventh {AAAI} Conference on Artificial Intelligence, {AAAI} 2023, Thirty-Fifth Conference on Innovative Applications of Artificial Intelligence, {IAAI} 2023, Thirteenth Symposium on Educational Advances in Artificial Intelligence, {EAAI} 2023, Washington, DC, USA, February 7-14, 2023}, 10444--10452. {AAAI} Press.

\bibitem[{Xiao, Wan, and Devadas(2023)}]{DBLP:conf/ccs/Xiao0D23}
Xiao, H.; Wan, J.; and Devadas, S. 2023.
\newblock Geometry of Sensitivity: Twice Sampling and Hybrid Clipping in Differential Privacy with Optimal Gaussian Noise and Application to Deep Learning.
\newblock In Meng, W.; Jensen, C.~D.; Cremers, C.; and Kirda, E., eds., \emph{Proceedings of the 2023 {ACM} {SIGSAC} Conference on Computer and Communications Security, {CCS} 2023, Copenhagen, Denmark, November 26-30, 2023}, 2636--2650. {ACM}.

\bibitem[{Xiao et~al.(2023)Xiao, Xiang, Wang, and Devadas}]{DBLP:conf/sp/XiaoXWD23}
Xiao, H.; Xiang, Z.; Wang, D.; and Devadas, S. 2023.
\newblock A Theory to Instruct Differentially-Private Learning via Clipping Bias Reduction.
\newblock In \emph{44th {IEEE} Symposium on Security and Privacy, {SP} 2023, San Francisco, CA, USA, May 21-25, 2023}, 2170--2189. {IEEE}.

\bibitem[{Xue et~al.(2024)Xue, Xue, Zhu, Luo, Zhang, Sun, and Lu}]{10262055}
Xue, R.; Xue, K.; Zhu, B.; Luo, X.; Zhang, T.; Sun, Q.; and Lu, J. 2024.
\newblock Differentially Private Federated Learning With an Adaptive Noise Mechanism.
\newblock \emph{IEEE Transactions on Information Forensics and Security}, 19: 74--87.

\bibitem[{Yang et~al.(2020)Yang, Tan, Zheng, Chen, and Yang}]{Yang2020}
Yang, L.; Tan, B.; Zheng, V.~W.; Chen, K.; and Yang, Q. 2020.
\newblock \emph{Federated Recommendation Systems}, 225--239.
\newblock Cham: Springer International Publishing.
\newblock ISBN 978-3-030-63076-8.

\bibitem[{Yang et~al.(2019)Yang, Liu, Chen, and Tong}]{10.1145/3298981}
Yang, Q.; Liu, Y.; Chen, T.; and Tong, Y. 2019.
\newblock Federated Machine Learning: Concept and Applications.
\newblock \emph{ACM Trans. Intell. Syst. Technol.}, 10(2).

\bibitem[{Yeom et~al.(2018)Yeom, Giacomelli, Fredrikson, and Jha}]{Yeom_Giacomelli_Fredrikson_Jha_2018}
Yeom, S.; Giacomelli, I.; Fredrikson, M.; and Jha, S. 2018.
\newblock Privacy Risk in Machine Learning: Analyzing the Connection to Overfitting.
\newblock In \emph{2018 IEEE 31st Computer Security Foundations Symposium (CSF)}.

\bibitem[{Zhang et~al.(2022)Zhang, Chen, Hong, Wu, and Yi}]{DBLP:conf/icml/ZhangCH0Y22}
Zhang, X.; Chen, X.; Hong, M.; Wu, S.; and Yi, J. 2022.
\newblock Understanding Clipping for Federated Learning: Convergence and Client-Level Differential Privacy.
\newblock In Chaudhuri, K.; Jegelka, S.; Song, L.; Szepesv{\'{a}}ri, C.; Niu, G.; and Sabato, S., eds., \emph{International Conference on Machine Learning, {ICML} 2022, 17-23 July 2022, Baltimore, Maryland, {USA}}, volume 162 of \emph{Proceedings of Machine Learning Research}, 26048--26067. {PMLR}.

\bibitem[{Zhu et~al.(2024)Zhu, Liu, Tang, and Niu}]{10255314}
Zhu, G.; Liu, X.; Tang, S.; and Niu, J. 2024.
\newblock Aligning Before Aggregating: Enabling Communication Efficient Cross-Domain Federated Learning via Consistent Feature Extraction.
\newblock \emph{IEEE Transactions on Mobile Computing}, 23(5): 5880--5896.

\end{thebibliography}

\end{document}